# MGCVAE: Multi-objective Inverse Design via Molecular Graph Conditional Variational Autoencoder

Myeonghun Lee[1] and Kyoungmin Min[2,*]

[1]School of Systems Biomedical Science, [2]School of Mechanical Engineering, Soongsil University, 369 Sangdo-ro, Dongjak-gu, Seoul 06978, Republic of Korea

## ABSTRACT

The ultimate goal of various fields is to directly generate molecules with desired properties, such as finding water-soluble molecules in drug development and finding molecules suitable for organic light-emitting diode (OLED) or photosensitizers in the field of development of new organic materials. In this respect, this study proposes a molecular graph generative model based on the autoencoder for de novo design. The performance of molecular graph conditional variational autoencoder (MGCVAE) for generating molecules having specific desired properties is investigated by comparing it to molecular graph variational autoencoder (MGVAE). Furthermore, multi-objective optimization for MGCVAE was applied to satisfy two selected properties simultaneously. In this study, two physical properties- $logP$ and molar refractivity were used as optimization targets for the purpose of designing de novo molecules, especially in drug discovery. As a result, it was confirmed that among generated molecules, 25.89% optimized molecules were generated in MGCVAE compared to 0.66% in MGVAE. Hence, it demonstrates that MGCVAE effectively produced drug-like molecules with two target properties. The results of this study suggest that these graph-based data-driven models are one of the effective methods of designing new molecules that fulfill various physical properties, such as drug discovery.

## Keywords

*Corresponding author: kmin.min@ssu.ac.kr (K. Min)

## INTRODUCTION

Finding new molecules with desired properties for de novo drug design is a significant and challenging task [1]. However, it is almost impossible to find these molecules in the high-dimensional chemical space of all molecules without prior knowledge. Recently, computational biologists have extensively integrated machine learning and deep learning into drug design and discovery processes [2], [3]. Primarily, the computational modeling along with those algorithms provides a promising way to identify and validate drug toxicity and drug monitoring of compounds [4].

The computer-based molecular design has received much attention as a solution to overcome experimental limitations [5]–[8]. High-throughput virtual screening can find molecules with desired physical properties through massive calculations with high accuracy and low cost. Such a method can aid in selecting appropriate molecules from the millions of molecules in the database, and then experiments can be followed for verifying them [9]. It is also interesting to note that the generative model, which recently emerged with the development of deep learning, encouraged application to new molecular designs that use generative models to propose new molecules likely to have ideal properties. Specifically, some successful long short-term memory (LSTM) network architectures have been demonstrated and validated for *in silico* screening by generating molecules with desired properties [10]–[12]. The computer-based molecular design has received much attention as a solution to overcome experimental limitations [5]–[8]. High-throughput virtual screening can find molecules with desired physical properties through massive calculations with high accuracy and low cost. Such a method can aid in selecting appropriate molecules from the millions of molecules in the database, and then experiments can be followed for verifying them [9]. It is also interesting to note that the generative model, which recently emerged with the development of deep learning, encouraged application to new molecular designs that use generative models to propose new molecules likely to have ideal properties. Specifically, some successful long short-term memory (LSTM) network architectures have been demonstrated and validated for *in silico* screening by generating molecules with desired properties [10]–[12].

Recurrent neural networks (RNN) still play an essential role for a molecular generation.

SMILES (Simplified Molecular-Input Line-Entry System) is a line notation for describing molecules using strings; thus, text generation models were developed [13]–[15]. In particular, RNN architectures based on LSTM achieve excellent results in natural language processing tasks, in which input is a token sequence of different lengths. Some implemented LSTM-based conditional variational autoencoder (CVAE) [16] or reinforcement learning strategies to generate molecules with specific properties [17], [18]. However, it is unfortunate that creating SMILES expressions of molecules often becomes unstable, resulting in a misconstruction. Generating a valid SMILES string requires the model to learn rules unrelated to molecular structure, such as SMILES grammar and atomic order, which adds unnecessary burden to the training process; Thus, making SMILES strings a less preferred representation [19]. Applying graphs to describe the molecules has been suggested to be more natural for data structures to overcome such issues. Compared to generating a grammatically valid SMILES string, graph-based molecular generation could be more efficient because the atoms constituting each node and the bonds constituting each edge are first determined so that all outputs are valid molecules after that [20], [21].

The graph-based generation models are generally presented with a sequential and an entire graph generation method. For example, some models produce final graphs, constructing additional nodes and edges [22]–[24]. Johnson *et al.* [25] proposed a sequential generative approach for graphs, and this framework is potentially applicable to molecular generation. JT-VAE [26] generates a tree of molecular fragments based on variational autoencoder (VAE) and then determines the final molecular graph through the subsequent assembly of the fragments, allowing the molecule to be progressively expanded while maintaining chemical validity at every step. In addition, MolecularRNN [27] uses the policy gradient algorithm to generate molecules with desired lipophilicity, drug-likeness, and melting point for drug development. Other examples include GCPN [28] based on reinforcement learning, assuming a sequential order of graph expansion steps, reported as models for generating molecules with desired properties. On the other hand, the method for generating whole graphs has been successfully applied to generating small molecule graphs. GraphVAE [29] using a graph convolution algorithm and MolGAN [30] based on reinforcement learning are representative models generating graphs simultaneously.

Optimization simultaneously of molecules concerning multiple properties remains a significant challenge [31]. Especially in the design of de novo molecules, for a molecule to act as a drug, certain conditions must be considered. In this study, as a case where multi-objective optimization is applied, the models were examined by generating molecules whether they satisfy the requirements for drug development. Since graph generation is still challenging to create large sizes, in this study, small molecules with less than 16 atoms (molecular weight < 200 g/mol) were generated; we note that compared to other generative models using QM9 [32], this is not a small number. In **Figure 1**, molecules with the desired log octanol-water partition coefficient (*logP*) and molar refractivity (*MR*) were generated using a molecular graph conditional variational autoencoder (MGCVAE). To verify the model performance, the results of a molecular graph variational autoencoder (MGVAE) without conditions and the results of MGCVAE using given specific conditions were compared. Finally, this study proposes and verifies the feasibility of generating molecules that satisfy two desired properties based on large amounts of data.

## METHOD

### Molecular graphs

The graph representation of molecules corresponds to atoms by nodes and bonds by edges, represented by an annotation matrix and an adjustment matrix. The annotation matrix ($N \times X$, $N$ is the number of atoms, $X$ is the number of types of atoms) represents each row as the one-hot encoding of atoms, and the adjacency matrix ($N \times N$) represents how each row and column corresponding to the atoms are binding. As shown in **Figure S1**, **Supporting Information (SI)**, the initial graph matrix of current models is reconstructed into the annotation matrix and the adjacency matrix to generate a complete molecular graph. The shape of the initial graph matrix is $(S, (1 + A + (S \times B)))$, where $S$ is the maximum number of atoms (the maximum graph size), $A$ is the number of atom types, and $B$ is the number of bond types.

The initial graph matrix is reconstructed into a molecular graph is illustrated in **Figure S1, SI**.

In **Figure S1**, the initial graph matrix for generating the molecular graph consists of the annotation matrix, and the adjacency matrix is transformed as follows: (1) "a" is the result of one-hot encoding of the number of atoms in the molecule, and the row below is not used based on this. Here it can have up to seven atoms, meaning that the molecule has six atoms. (2) "b" is an annotation matrix resulting from one-hot encoding for what each atom is. (3) "c-1" to "c-7" are one-hot encoding results for each row of the adjacency matrix. Therefore, it is reconstructed to form the annotation matrix and adjacency matrix, and it is converted into SMILES of the appropriate molecular graph.

**Conditional variational autoencoder (CVAE)**

The core of this study is applying multi-objective optimization based on graph description rather than string-based, and a data-driven autoencoder was implemented. Such an approach is dissimilar to the reinforcement learning-based generative model [32], [33] that predicts or calculates and rewards for the physical properties of molecules generated in the training process. This reinforcement learning-based optimization method is also powerful under the condition that it can give appropriate rewards when the properties of the generated molecules are measured or predicted. However, it is highly dependent which of predictive models is used. It is important to note that the autoencoder-based optimization method can overcome this limitation because it uses data and known physical properties.

For this reason, MGVAE and MGCVAE were implemented to generate molecules and compare their performance. As a molecular generative model, MGVAE produces molecules, which are physically (*logP* and *MR* in this study) similar to a given dataset. In the meanwhile MGCVAE is beneficial in producing molecules physically similar to a given dataset with a given specific condition. As shown in **Figure 1**, the model architecture of the current study starts with the encoder, which receives an initial graph matrix to generate an adjacency matrix and annotation matrix representing the molecule along with specific condition vectors. Then the latent space again receives the condition vectors. Finally, the decoder generates a graph of molecules with particular properties.

The significant difference between the two models is originated from different objective functions. The objective functions of each MGVAE and MGCVAE are as follows:

$$E[logP(X|z)] - D_{KL}[Q(z|X)||P(z)]$$

$$E[logP(X|z,c)] - D_{KL}[Q(z|X,c)||P(z|c)]$$

where $E$ represents the expectation value, $P$ and $Q$ indicate the probability distribution, represents the Kullback-Leibler divergence, $X$, $z$, and $c$ are the data, latent space, and condition vectors, respectively, and $P(X|z)$ represent encoder and decoder in autoencoder, respectively. Specifically, the formula for MGVAE is known as the Evidence Lower Bound (ELBO).

The main differences between these two models are originated from different objective functions according to condition vector $c$. In this study, the molecular properties to control corresponded as condition vectors, which are numerical one-hot encoded vectors. Therefore, the decoder of the trained model generates molecules with desired properties according to the given condition vector along with the latent vector. As a result, MGCVAE can generate molecules with target properties imposed by condition vectors.

**Dataset**

This study selected 1,363,452 molecules with 16 or fewer elements (nodes) obtained from the ZINC database [34], as shown in **Figure 2**. Molecules are composed of 12 types of atoms (B, C, N, O, F, Si, P, S, Cl, Br, Sn, and I) and four types of bonds (single, double, triple, and aromatic bonds). In addition, by RDKit [35] calculation methods, these molecules also have $logP$ between -6 -6 and 5, where $P$ is defined as the ratio of the concentrations between two solvents of a solute [36]. $MR$ between 5 and 95, which is an important criterion to measure the steric factor and related to the size and molecular weight of molecules [37]. In particular, in this study, molecules suitable for graph generation (e.g., SMILES without '+', '–', and '.') were collected from the ZINC database. This ZINC dataset was divided into a training set and a test set as a ratio of 9:1, and the training process was confirmed as shown in **Figure S2, SI**. The input initial graph matrix and output initial graph matrix of the autoencoder was trained in the direction of decreasing the loss so that they could be the same as possible. According to Lipinski's rule of five (RO5) [38], drug-like molecules have $logP$ between -0.4 and 5.6 and $MR$ between 40 and 130; hence it was investigated to generate molecules that have $logP$ between 0 and 3, and $MR$ between 20 and 60, including both a meaningful range as a drug and a range in which the model can be generated by the distribution of the given dataset.

## RESULTS AND DISCUSSIONS

### Multi-objective optimization

By training the distribution of $logP$ and $MR$ for molecules of selected ZINC dataset using MGVAE and MGCVAE, the $logP$ and $MR$ values of molecules generated by MGVAE and MGCVAE are obtained. The two models were trained with the same dataset, and 10,000 molecules were generated with each model to compare their performance. While MGVAE generated 10,000 cases without any constraints, MGCVAE generated molecules with the first condition ($logP$, $C_1 = \{0, 1, 2, 3\}$), and with the second condition ($MR$, $C_2 = \{20, 30, 40, 50, 60\}$). 10,000 molecules are generated from each condition, leading to a total of 200,000 molecules are suggested.

To optimize the two target properties, the performance was evaluated by checking how many generated molecules were satisfying the properties within the target range when the conditions within the corresponding range were respectively given. To confirm the performance of MGCVAE more clearly, molecules generated from MGVAE, in which no condition is applied, are counted whether they are matched the range of each condition. First of all, as shown in **Figure 3 and Table 1**, when $logP$ was close to 3 and $MR$ was close to 20, no molecules satisfying both conditions were generated; we note that even in the ZINC dataset, there was no data that satisfies both of these conditions. However, except for this, it was clearly validated that MGCVAE generated more optimized molecules than MGVAE in all conditions. For visualization purposes, the MGCVAE results of **Table 1** are converted into 3D in **Figure S3, SI**. Except for $C_1 = \{3\}$ and $C_2 = \{20\}$, when using MGCVAE, the molecular conditions were satisfied from a minimum of 23.46% (13.99% in MGVAE) to a maximum of 44.12% (30.90% in MGVAE) for *logP*, and from a minimum of 38.57% (0.11% in MGVAE) to a maximum of 82.72% (36.66% in MGVAE) for *MR*. In addition, the cases where both conditions were satisfied ranged a maximum of 32.78% (16.90% in MGVAE). It can be seen that this is an effective optimization compared to the case where the MGVAE satisfies both conditions at the same time from a minimum of 0.00% to a maximum of 16.90%. In addition, the difference in the degree of optimization between MGCVAE and MGVAE was 26.86% in *logP*, 58.30% in *MR*, and 25.23% in both, with MGCVAE being more effective. Such remarkable performance in MGCVAE suggests that it is suitable for generating new organic materials or drugs with two desired properties, such as *logP* and *MR*.

In this study, the performance of the single-objective optimization was not solely verified. Still, the degree of optimization of a single property could be confirmed from the results on the multi-objective optimization. Thus, among the cases where $C_1 = \{0, 1, 2, 3\}$ and $C_2 = \{20, 30, 40, 50, 60\}$, in the case of $C_1 = \{0\}$ and $C_2 = \{20, 30, 40, 50, 60\}$, the generated molecules were checked together to confirm how much $C_1 = \{0\}$ was optimized. As a result, as shown in **Figure S4, SI,** a maximum of 38.89% at $C_1 = \{2\}$ and a maximum of 78.65% at $C_2 = \{60\}$, better results were confirmed in *MR* than *logP*.

**Molecular library similarity**

Comparing the chemical space between the generated molecules and the molecules in the dataset is a way to visually compare how chemically similar the generated molecules are to the dataset and is often attempted [39]. This comparison shows how the two sets of molecules cover similar chemical spaces. For this, molecular fingerprints and dimensionality reduction algorithms were used to visually analyze the similarity between the generated molecules and the molecules of the training set, as shown in **Figure 4**. Thus, MACCS molecular fingerprints (167-bits) [40] and Morgan molecular fingerprints (1024-bits) [41], which are typically used for in silico screening, were implemented for randomly selected 50,000 molecules from ZINC dataset, 5,000 molecules generated from MGVAE, 50,000 molecules out of all molecules generated for all conditions of MGCVAE, respectively. Each molecular fingerprint was reduced and visualized in two dimensions using principal component analysis (PCA). In the case of MGVAE, the generated molecules were located relative to the left, but overlapped in the clusters formed by the molecules in the ZINC dataset. In particular, in the case of MGCVAE, the results and distribution of ZINC were different because the molecules generated according to the conditions were different. For example, MGVAE was almost similar to the histogram of ZINC, whereas in MGCVAE, the positions and distributions of the formed peaks were different. As a result, the positions and clusters of each generated molecule by MGVAE and the positions of bright areas appeared similar to those of the molecules from the ZINC dataset, but the molecules generated by MGCVAE showed different distributions in the positions of the corresponding clusters. This suggests that both generative models generate molecules similar to the existing datasets, especially in MGCVAE, which allows more molecules to be generated at specific locations.

**Model performance evaluation**

It is imperative to check the following metrics for the generative model because the chemical space is limited when the model trained on a particular dataset generates molecules. Therefore, to evaluate the performance of MGVAE, validity (the ratio between the number of valid and all generated molecules), novelty (the ratio between the set of valid samples that are not in the dataset and the total number of valid samples), and uniqueness (the ratio between the number of unique samples and valid samples and it measures the degree of variety during sampling) were measured together. As shown in **Table 2**, the results were confirmed with full marks for validity, novelty, and uniqueness. In the case of uniqueness, it is remarkable because it is an indicator that did not reach a perfect score in previous studies. Furthermore, in the case of MGCVAE, metrics were also measured for $C_1 = \{3\}$ and $C_2 = \{60\}$, which are the conditions that generate the most optimized molecules, and the results showed that only the uniqueness was slightly lower than that of MGVAE. Example molecules from ZINC dataset and generated by MGCVAE are randomly chosen and shown in **Figure S5, SI**.

Finally, in this study, the performance of MGVAE and MGCVAE, which are proposed for the purpose of multi-objective optimization, were compared and analyzed according to particular conditions. The results confirmed that desired molecules were well generated in MGCVAE. In addition, the molecular fingerprints of the generated molecules were visualized to verify that they were similar to the dataset, and the model was evaluated through three evaluation metrics.

## CONCLUSION

In this study, MGVAE and MGCVAE were proposed as graph-based deep learning molecular generation models restructured from the initial graph matrix. In addition, in the case of MGCVAE, multi-objective optimization was performed for drug development. To evaluate this, results were compared with MGVAE, and molecules with the desired target properties (*logP* and *MR*) were directly generated by simultaneously controlling multiple target properties by assigning them to condition vectors in this model. As a result, when *logP* is close to 3 and *MR* is close to 60, it was confirmed that MGCVAE generated 32.78% of molecules when MGVAE generated 9.01% of molecules, and the chemical spaces of the generated molecules and the latent spaces of the models were visually examined. Therefore, the models in this study have the potential to be applied to efficient de novo drug design. However, improvements and future work in this study include generating and validating large graphs of 20 nodes or more with sizes suitable for drug-like molecules and optimizing three or more properties.

## ACKNOWLEDGMENT

This research was supported by the MSIT(Ministry of Science, ICT), Korea, under the High-Potential Individuals Global Training Program)(2021-0-01539) supervised by the IITP(Institute for Information & Communications Technology Planning & Evaluation). This work was supported by the National Research Foundation of Korea (NRF) grant funded by

## DATA AVAILABILITY

The dataset and the main source code used in this study are available at the GitHub page https://github.com/mhlee216/MGCVAE. All other relevant data are available from the authors upon reasonable request.

## APPENDIX A. SUPPORTING INFORMATION

## REFERENCES

[1] G. Schneider and U. Fechner, "Computer-based de novo design of drug-like molecules," *Nature Reviews Drug Discovery 2005 4:8*, vol. 4, no. 8, pp. 649–663, Aug. 2005, doi: 10.1038/nrd1799.

[2] M. Wainberg, D. Merico, A. Delong, and B. J. Frey, "Deep learning in biomedicine," *Nature Biotechnology 2018 36:9*, vol. 36, no. 9, pp. 829–838, Oct. 2018, doi: 10.1038/nbt.4233.

[3] P. Mamoshina, A. Vieira, E. Putin, and A. Zhavoronkov, "Applications of Deep Learning in Biomedicine," *Molecular Pharmaceutics*, vol. 13, no. 5, pp. 1445–1454, May 2016, doi: 10.1021/ACS.MOLPHARMACEUT.5B00982/SUPPL_FILE/MP5B00982_SI_001.PDF.

[4] R. Gupta, D. Srivastava, M. Sahu, S. Tiwari, R. K. Ambasta, and P. Kumar, "Artificial intelligence to deep learning: machine intelligence approach for drug discovery," *Molecular Diversity 2021 25:3*, vol. 25, no. 3, pp. 1315–1360, Apr. 2021, doi: 10.1007/S11030-021-10217-3.

[5] B. K. Shoichet, "Virtual screening of chemical libraries," *Nature 2004 432:7019*, vol. 432, no. 7019, pp. 862–865, Dec. 2004, doi: 10.1038/nature03197.

[6] T. Scior *et al.*, "Recognizing Pitfalls in Virtual Screening: A Critical Review," *Journal of Chemical Information and Modeling*, vol. 52, no. 4, pp. 867–881, Apr. 2012, doi: 10.1021/CI200528D.

[7] T. Cheng, Q. Li, Z. Zhou, Y. Wang, and S. H. Bryant, "Structure-Based Virtual Screening for Drug Discovery: a Problem-Centric Review," *The AAPS Journal*, vol. 14, no. 1, p. 133, Mar. 2012, doi: 10.1208/S12248-012-9322-0.

[8] J. L. Reymond, R. van Deursen, L. C. Blum, and L. Ruddigkeit, "Chemical space as a source for new drugs," *MedChemComm*, vol. 1, no. 1, pp. 30–38, Jul. 2010, doi: 10.1039/C0MD00020E.

[9]     J. Lim, S. Ryu, J. W. Kim, and W. Y. Kim, "Molecular generative model based on conditional variational autoencoder for de novo molecular design," *Journal of Cheminformatics*, vol. 10, no. 1, pp. 1–9, Dec. 2018, doi: 10.1186/S13321-018-0286-7/TABLES/2.

[10]    M. H. S. Segler, T. Kogej, C. Tyrchan, and M. P. Waller, "Generating focused molecule libraries for drug discovery with recurrent neural networks," *ACS Central Science*, vol. 4, no. 1, pp. 120–131, Jan. 2018, doi: 10.1021/ACSCENTSCI.7B00512/SUPPL_FILE/OC7B00512_SI_002.ZIP.

[11]    R. Winter, F. Montanari, F. Noé, and D. A. Clevert, "Learning continuous and data-driven molecular descriptors by translating equivalent chemical representations," *Chemical Science*, vol. 10, no. 6, pp. 1692–1701, Feb. 2019, doi: 10.1039/C8SC04175J.

[12]    A. Gupta, A. T. Müller, B. J. H. Huisman, J. A. Fuchs, P. Schneider, and G. Schneider, "Generative Recurrent Networks for De Novo Drug Design," *Molecular Informatics*, vol. 37, no. 1–2, Jan. 2018, doi: 10.1002/MINF.201700111.

[13]    M. H. S. Segler, T. Kogej, C. Tyrchan, and M. P. Waller, "Generating focused molecule libraries for drug discovery with recurrent neural networks," *ACS Central Science*, vol. 4, no. 1, pp. 120–131, Jan. 2018, doi: 10.1021/ACSCENTSCI.7B00512/SUPPL_FILE/OC7B00512_SI_002.ZIP.

[14]    A. Gupta, A. T. Müller, B. J. H. Huisman, J. A. Fuchs, P. Schneider, and G. Schneider, "Generative Recurrent Networks for De Novo Drug Design," *Molecular Informatics*, vol. 37, no. 1–2, Jan. 2018, doi: 10.1002/MINF.201700111.

[15]    E. J. Bjerrum and B. Sattarov, "Improving Chemical Autoencoder Latent Space and Molecular De Novo Generation Diversity with Heteroencoders," *Biomolecules*, vol. 8, no. 4, Dec. 2018, doi: 10.3390/BIOM8040131.

[16]    J. Lim, S. Ryu, J. W. Kim, and W. Y. Kim, "Molecular generative model based on conditional variational autoencoder for de novo molecular design," *Journal of Cheminformatics*, vol. 10, no. 1, pp. 1–9, Dec. 2018, doi: 10.1186/S13321-018-0286-7/TABLES/2.

[17]    M. Popova, O. Isayev, and A. Tropsha, "Deep reinforcement learning for de novo drug design," *Science Advances*, vol. 4, no. 7, p. 7885, Jul. 2018, doi: 10.1126/SCIADV.AAP7885.

[18]    G. Guimaraes, B. Sanchez-Lengeling, C. Outeiral, P. Luis, C. Farias, and A. Aspuru-Guzik, "Objective-Reinforced Generative Adversarial Networks (ORGAN) for Sequence Generation Models," May 2017, Accessed: Feb. 06, 2022. [Online]. Available: https://arxiv.org/abs/1705.10843v3

[19]    Y. Li, L. Zhang, and Z. Liu, "Multi-Objective De Novo Drug Design with Conditional Graph Generative Model," *Journal of Cheminformatics*, vol. 10, no. 1, Jan. 2018, doi: 10.1186/s13321-018-0287-6.

[20]    Ł. Maziarka, A. Pocha, J. Kaczmarczyk, K. Rataj, T. Danel, and M. Warchoł, "Mol-CycleGAN: A generative model for molecular optimization," *Journal of Cheminformatics*, vol. 12, no. 1, pp. 1–18, Jan. 2020, doi: 10.1186/S13321-019-0404-1/FIGURES/22.

[21]    R. Mercado *et al.*, "Graph networks for molecular design," *Machine Learning: Science and Technology*, vol. 2, no. 2, p. 025023, Mar. 2021, doi: 10.1088/2632-2153/ABCF91.

[22]    Y. Li, O. Vinyals, C. Dyer, R. Pascanu, and P. Battaglia, "Learning Deep Generative Models of Graphs," Mar. 2018, Accessed: Feb. 06, 2022. [Online]. Available: https://arxiv.org/abs/1803.03324v1

[23]    G. N. C. Simm and J. M. Hernández-Lobato, "A Generative Model for Molecular Distance Geometry," *37th International Conference on Machine Learning, ICML 2020*, vol. PartF168147-12, pp. 8896–8905, Sep. 2019, Accessed: Feb. 06, 2022. [Online]. Available: https://arxiv.org/abs/1909.11459v4

[24]    Y. Li, L. Zhang, and Z. Liu, "Multi-objective de novo drug design with conditional graph generative model," *Journal of Cheminformatics*, vol. 10, no. 1, pp. 1–24, Dec. 2018, doi: 10.1186/S13321-018-0287-6/FIGURES/14.

[25]    C. J. Maddison, A. Mnih, and Y. W. Teh, "Learning Graphical State Transitions," *5th International Conference on Learning Representations, ICLR 2017 - Conference Track Proceedings*, Oct. 2016.

[26]    W. Jin, R. Barzilay, and T. Jaakkola, "Junction Tree Variational Autoencoder for Molecular Graph Generation," *35th International Conference on Machine Learning, ICML 2018*, vol. 5, pp. 3632–3648, Feb. 2018, Accessed: Feb. 06, 2022. [Online]. Available: https://arxiv.org/abs/1802.04364v4

[27]    M. Popova, M. Shvets, J. Oliva, and O. Isayev, "MolecularRNN: Generating realistic molecular graphs with optimized properties," May 2019, Accessed: Feb. 06, 2022. [Online]. Available: https://arxiv.org/abs/1905.13372v1

[28]    J. You, B. Liu, R. Ying, V. Pande, and J. Leskovec, "Graph Convolutional Policy Network for Goal-Directed Molecular Graph Generation," *Advances in Neural Information Processing Systems*, vol. 2018-December, pp. 6410–6421, Jun. 2018, Accessed: Feb. 06, 2022. [Online]. Available: https://arxiv.org/abs/1806.02473v3

[29]    M. Simonovsky and N. Komodakis, "GraphVAE: Towards Generation of Small Graphs Using Variational Autoencoders," *Lecture Notes in Computer Science (including subseries Lecture Notes in Artificial Intelligence and Lecture Notes in Bioinformatics)*, vol. 11139 LNCS, pp. 412–422, Feb. 2018, doi: 10.1007/978-3-030-01418-6_41.

[30]    N. de Cao and T. Kipf, "MolGAN: An implicit generative model for small molecular graphs," May 2018, Accessed: Feb. 06, 2022. [Online]. Available: https://arxiv.org/abs/1805.11973v1

[31]    K. M. Honorio, T. L. Moda, and A. D. Andricopulo, "Pharmacokinetic Properties and In Silico ADME Modeling in Drug Discovery," *Medicinal Chemistry*, vol. 9, no. 2, pp. 163–176, Jan. 2013, doi: 10.2174/1573406411309020002.

[32]    N. de Cao and T. Kipf, "MolGAN: An implicit generative model for small molecular graphs," May 2018, Accessed: Feb. 06, 2022. [Online]. Available: https://arxiv.org/abs/1805.11973v1

[33]    M. Popova, O. Isayev, and A. Tropsha, "Deep reinforcement learning for de novo drug design,"

*Science Advances*, vol. 4, no. 7, p. 7885, Jul. 2018, doi: 10.1126/SCIADV.AAP7885.

[34] J. J. Irwin and B. K. Shoichet, "ZINC – A Free Database of Commercially Available Compounds for Virtual Screening," *Journal of chemical information and modeling*, vol. 45, no. 1, p. 177, Jan. 2005, doi: 10.1021/CI049714.

[35] "RDKit." https://www.rdkit.org/ (accessed Feb. 06, 2022).

[36] J. Comer and K. Tam, "Lipophilicity Profiles: Theory and Measurement," *Pharmacokinetic Optimization in Drug Research*, pp. 275–304, Jun. 2007, doi: 10.1002/9783906390437.CH17.

[37] Z. Almi, S. Belaidi, T. Lanez, and N. Tchouar, "Structure Activity Relationships, QSAR Modeling and Drug-Like Calculations of TP Inhibition of 1,3,4-Oxadiazoline-2-Thione Derivatives," *International Letters of Chemistry, Physics and Astronomy*, vol. 37, pp. 113–124, Aug. 2014, doi: 10.18052/WWW.SCIPRESS.COM/ILCPA.37.113.

[38] C. A. Lipinski, F. Lombardo, B. W. Dominy, and P. J. Feeney, "Experimental and computational approaches to estimate solubility and permeability in drug discovery and development settings," *Advanced Drug Delivery Reviews*, vol. 23, no. 1–3, pp. 3–25, Jan. 1997, doi: 10.1016/S0169-409X(96)00423-1.

[39] O. Prykhodko *et al.*, "A de novo molecular generation method using latent vector based generative adversarial network," *Journal of Cheminformatics*, vol. 11, no. 1, pp. 1–13, Dec. 2019, doi: 10.1186/S13321-019-0397-9/FIGURES/9.

[40] J. L. Durant, B. A. Leland, D. R. Henry, and J. G. Nourse, "Reoptimization of MDL Keys for Use in Drug Discovery," *Journal of Chemical Information and Computer Sciences*, vol. 42, no. 6, pp. 1273–1280, Nov. 2002, doi: 10.1021/CI010132R.

[41] D. Rogers and M. Hahn, "Extended-Connectivity Fingerprints," *Journal of Chemical Information and Modeling*, vol. 50, no. 5, pp. 742–754, May 2010, doi: 10.1021/CI100050T.

[42] M. J. Kusner, B. Paige, and J. M. Hernández-Lobato, "Grammar Variational Autoencoder." PMLR, pp. 1945–1954, Jul. 17, 2017. Accessed: Feb. 06, 2022. [Online]. Available: https://proceedings.mlr.press/v70/kusner17a.html

[43] R. Gómez-Bombarelli *et al.*, "Automatic Chemical Design Using a Data-Driven Continuous Representation of Molecules," *ACS Central Science*, vol. 4, no. 2, pp. 268–276, Feb. 2018, doi: 10.1021/ACSCENTSCI.7B00572/SUPPL_FILE/OC7B00572_LIVESLIDES.MP4.

[44] M. Simonovsky and N. Komodakis, "GraphVAE: Towards Generation of Small Graphs Using Variational Autoencoders," *Lecture Notes in Computer Science (including subseries Lecture Notes in Artificial Intelligence and Lecture Notes in Bioinformatics)*, vol. 11139 LNCS, pp. 412–422, Oct. 2018, doi: 10.1007/978-3-030-01418-6_41.

[45] Q. Liu, M. Allamanis, M. Brockschmidt, and A. L. Gaunt, "Constrained Graph Variational Autoencoders for Molecule Design," *Advances in Neural Information Processing Systems*, vol. 2018-December, pp. 7795–7804, May 2018, Accessed: Feb. 06, 2022. [Online]. Available:

https://arxiv.org/abs/1805.09076v2

[46]    P. Bongini, M. Bianchini, and F. Scarselli, "Molecular generative Graph Neural Networks for Drug Discovery," *Neurocomputing*, vol. 450, pp. 242–252, Aug. 2021, doi: 10.1016/J.NEUCOM.2021.04.039.

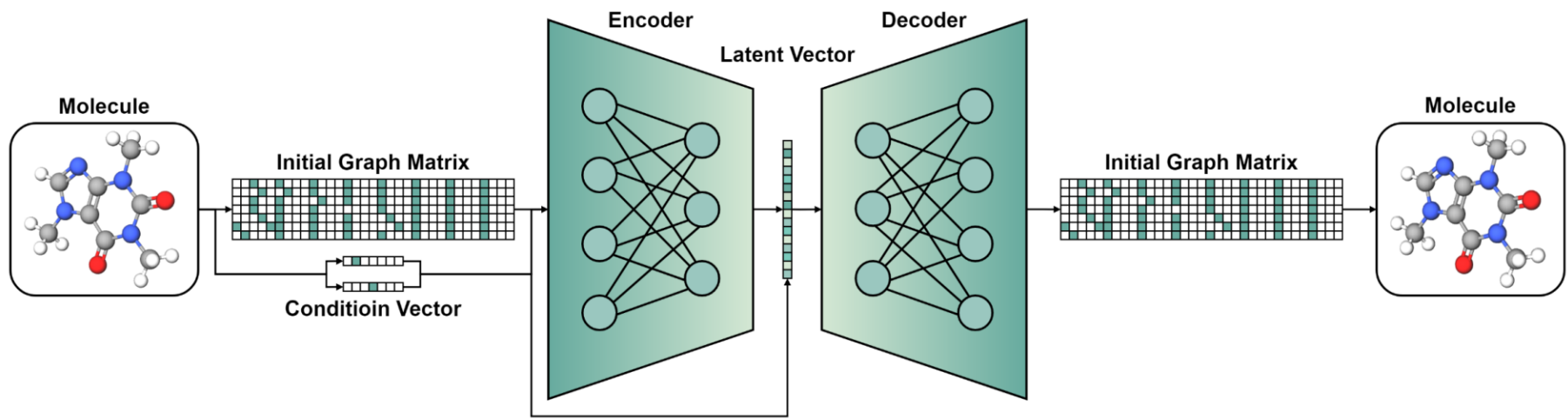

**Figure 1.** Architecture of the MGCVAE model.

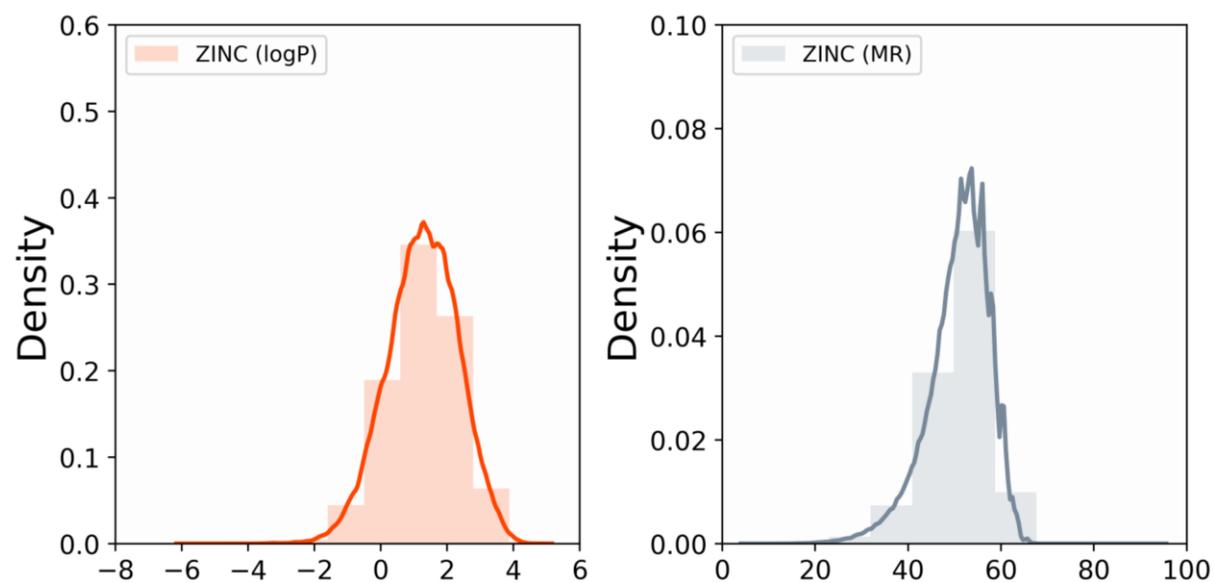

| Dataset | Molecules | Atoms | Atom Types | Bond Types | logP | MR |
|---------|-----------|-------|------------|------------|------|-----|
| ZINC | 1,363,452 | 16 | 12 | 4 | $-6 < logP < 5$ | $5 < MR < 95$ |

**Figure 2.** Histograms of *logP* and *MR* of the molecules and related information of the molecules included in the ZINC dataset used in this study.

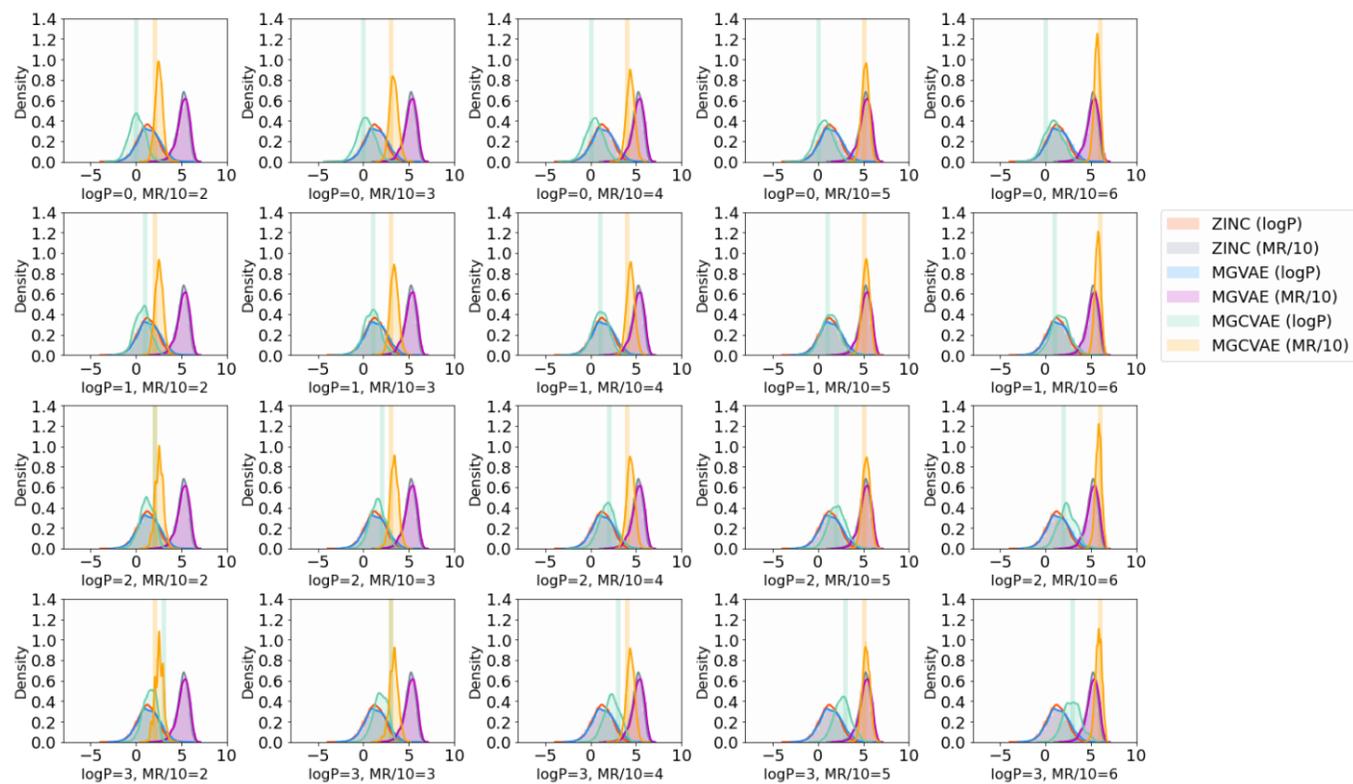

**Figure 3.** Comparison of multi-objective optimization results using MGCVAE, ZINC dataset, and MGVAE results in this study. Detailed figures are shown in **Table 1**.

| MGCVAE | | | MGVAE | | | Difference between MGCVAE and MGVAE | | | Condition | |
|---|---|---|---|---|---|---|---|---|---|---|
| logP (%) | MR (%) | Both (%) | logP (%) | MR (%) | Both (%) | logP (%) | MR (%) | Both (%) | logP | MR |
| 44.12 | 47.23 | 21.73 | 17.26 | 0.11 | 0.08 | 26.86 | 47.12 | 21.65 | 0 | 20 |
| 43.38 | 45.32 | 18.73 | 30.90 | 0.11 | 0.02 | 12.48 | 45.21 | 18.71 | 1 | 20 |
| 34.14 | 39.93 | 8.73 | 28.73 | 0.11 | 0.00 | 5.41 | 39.82 | 8.73 | 2 | 20 |
| 5.14 | 38.57 | 0.00 | 13.99 | 0.11 | 0.00 | -8.85 | 38.46 | 0.00 | 3 | 20 |
| 40.66 | 59.60 | 25.89 | 17.26 | 1.72 | 0.66 | 23.40 | 57.88 | 25.23 | 0 | 30 |
| 41.59 | 57.07 | 24.19 | 30.90 | 1.72 | 0.62 | 10.69 | 55.35 | 23.57 | 1 | 30 |
| 40.05 | 56.98 | 20.12 | 28.73 | 1.72 | 0.15 | 11.32 | 55.26 | 19.97 | 2 | 30 |
| 23.46 | 60.02 | 11.58 | 13.99 | 1.72 | 0.06 | 9.47 | 58.30 | 11.52 | 3 | 30 |
| 36.39 | 58.25 | 22.84 | 17.26 | 12.50 | 3.83 | 19.13 | 45.75 | 19.01 | 0 | 40 |
| 41.26 | 56.82 | 25.85 | 30.90 | 12.50 | 4.40 | 10.36 | 44.32 | 21.45 | 1 | 40 |
| 42.31 | 55.59 | 23.18 | 28.73 | 12.50 | 2.30 | 13.58 | 43.09 | 20.88 | 2 | 40 |
| 33.88 | 56.10 | 16.92 | 13.99 | 12.50 | 0.57 | 19.89 | 43.60 | 16.35 | 3 | 40 |
| 32.43 | 72.22 | 26.71 | 17.26 | 48.59 | 10.44 | 15.17 | 23.63 | 16.27 | 0 | 50 |
| 36.61 | 67.04 | 27.81 | 30.90 | 48.59 | 16.90 | 5.71 | 18.45 | 10.91 | 1 | 50 |
| 40.68 | 63.90 | 26.97 | 28.73 | 48.59 | 13.31 | 11.95 | 15.31 | 13.66 | 2 | 50 |
| 40.78 | 65.94 | 26.80 | 13.99 | 48.59 | 4.36 | 26.79 | 17.35 | 22.44 | 3 | 50 |
| 26.85 | 72.78 | 15.54 | 17.26 | 36.66 | 2.25 | 9.59 | 36.12 | 13.29 | 0 | 60 |
| 33.11 | 78.98 | 24.24 | 30.90 | 36.66 | 8.97 | 2.21 | 42.32 | 15.27 | 1 | 60 |
| 37.62 | 82.72 | 31.58 | 28.73 | 36.66 | 12.97 | 8.89 | 46.06 | 18.61 | 2 | 60 |
| 38.08 | 80.72 | 32.78 | 13.99 | 36.66 | 9.01 | 24.09 | 44.06 | 23.77 | 3 | 60 |

**Table 1**. The multi-objective optimization results. The results of MGCVAE and MGVAE are displayed along with each property and the result that satisfies both. Each result is the percentage of the number of molecules that, rounded up, have the same value as the given condition.

**(a)**

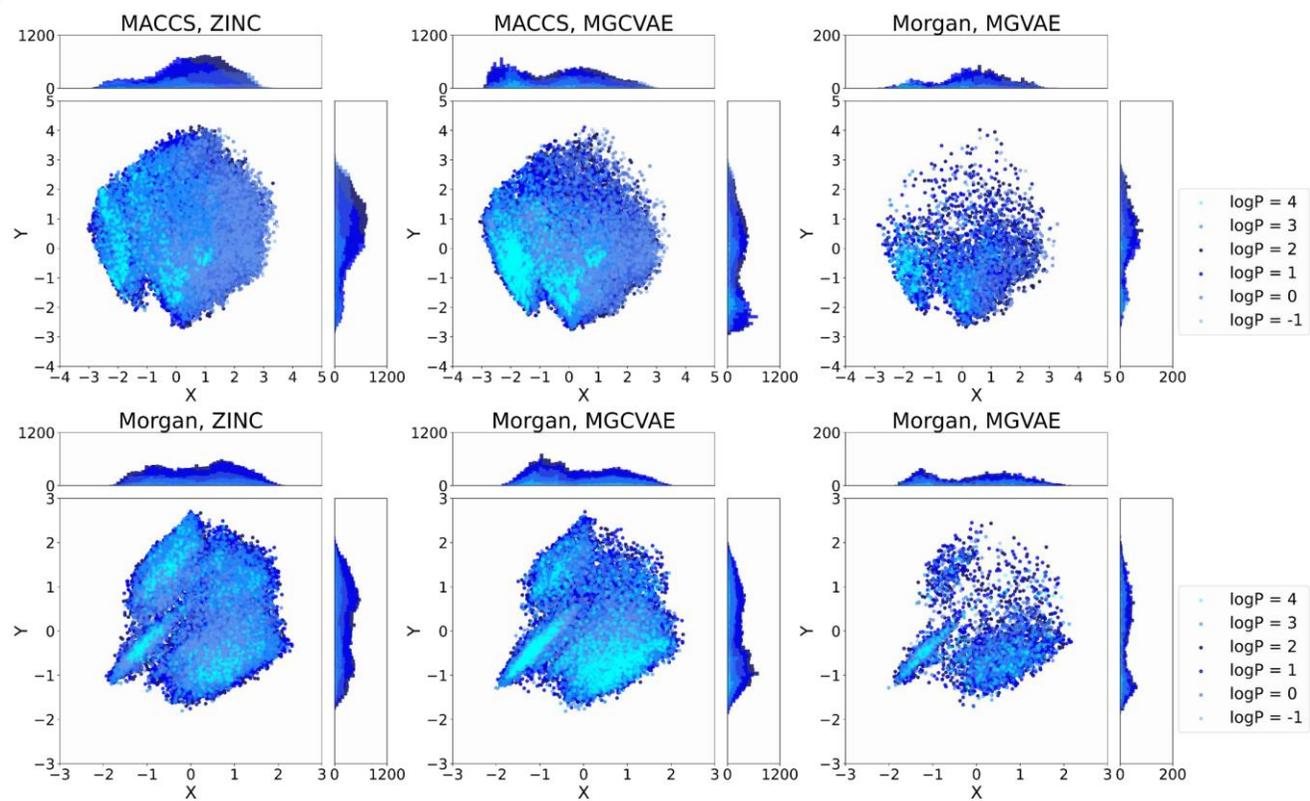

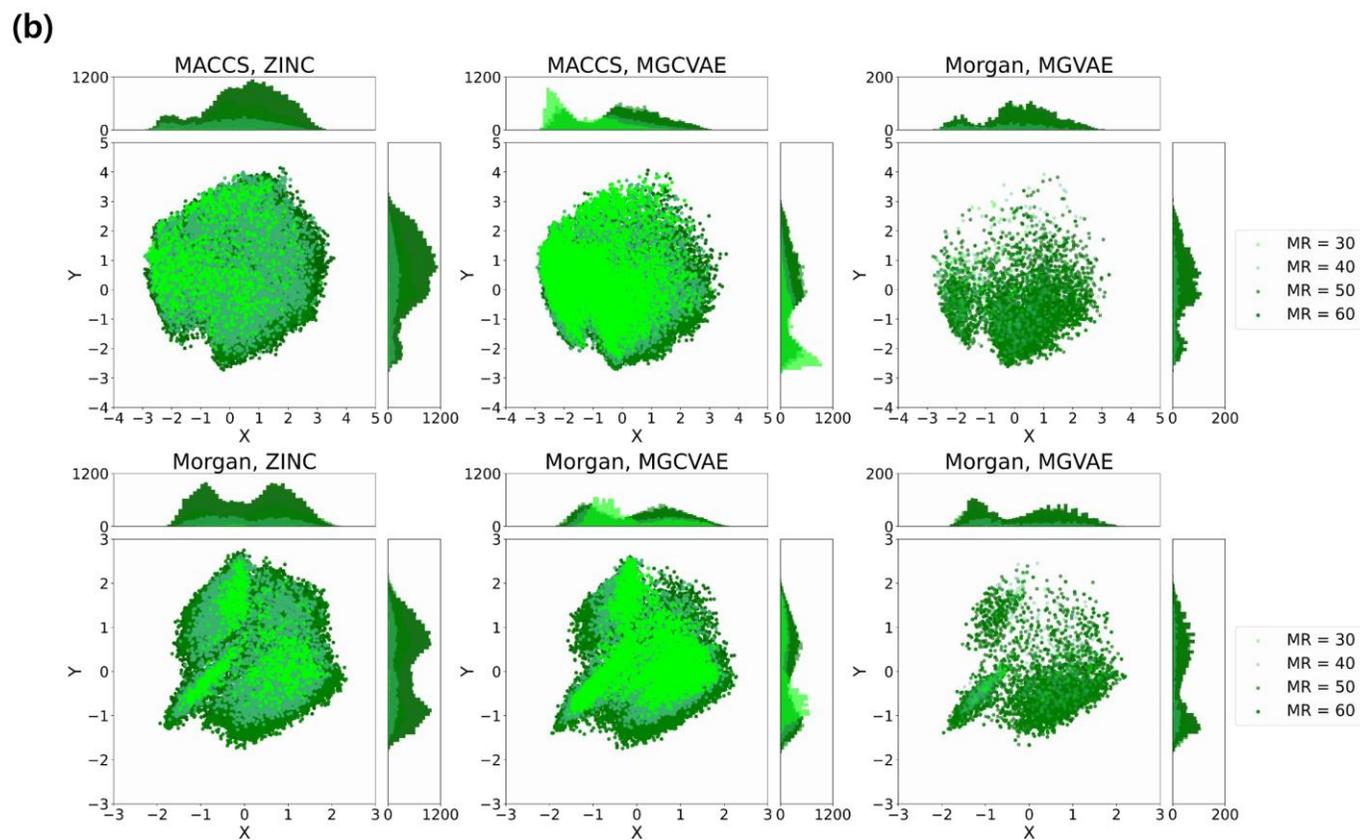

**Figure 4.** A two-dimensional reduced visualization using PCA of the MACCS and Morgan molecular fingerprints of the ZINC dataset and generated molecules, respectively. Each property is shown after rounding, and based on ZINC, the larger the number of data, the darker the color is drawn on the back side.

| Model | Valid | Unique | Novel |
|-------|-------|--------|-------|
| GrammarVAE [42] | 0.310 | 0.108 | **1.000** |
| ChemVAE [43] | 0.170 | 0.310 | 0.980 |
| GraphVAE [44] | 0.140 | 0.316 | **1.000** |
| CGVAE [45] | **1.000** | 0.998 | **1.000** |
| MG$^2$N$^2$ [46] | 0.753 | 0.107 | **1.000** |
| **MGVAE** | **1.000** | **1.000** | **1.000** |
| **MGCVAE** | **1.000** | 0.998 | **1.000** |

**Table 2.** Validity, Uniqueness, and Novelty of generated molecules assessing the quality of GCVAE, MGCVAE and the baselines on the ZINC dataset.

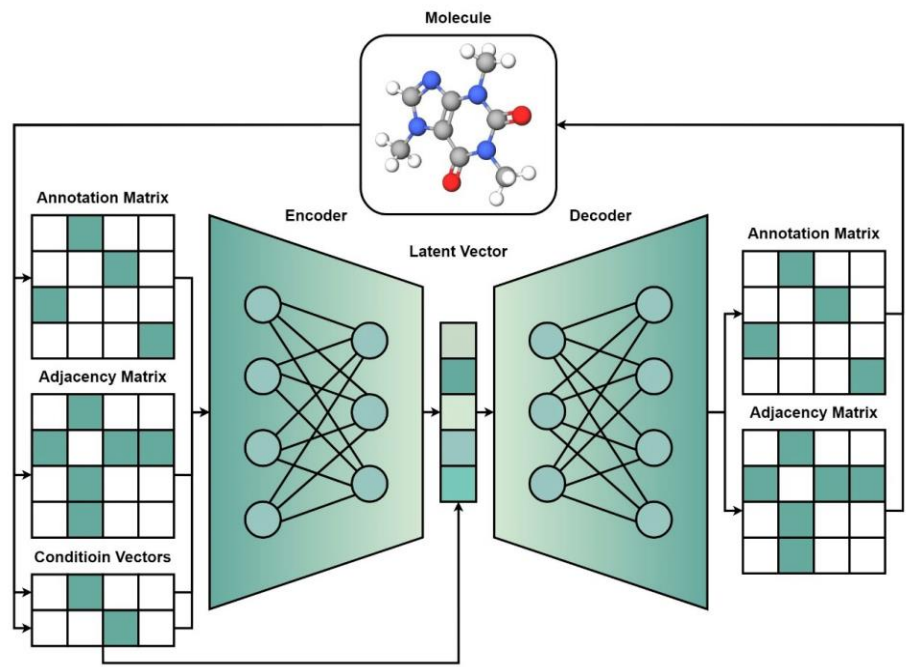

**TOC**